\documentclass{article}
\usepackage{spconf,amsmath,graphicx,hyperref}
\usepackage{graphicx}
\usepackage{xcolor}
\usepackage[table,xcdraw]{xcolor}
\usepackage{subcaption}
\usepackage{cite}
\usepackage{todonotes}
\usepackage{url}
\usepackage{multirow}
\usepackage[utf8]{inputenc}
\DeclareUnicodeCharacter{2212}{-}

\title{Mispronunciation Detection and Diagnosis Without Model Training: A Retrieval-Based Approach}

\name{
\begin{tabular}{c}
Huu Tuong Tu$^{1,2}$ \quad Ha Viet Khanh$^1$ \quad Tran Tien Dat$^1$ \quad
Vu Huan$^{3}$ \quad Thien Van Luong$^{3}$ \\ Nguyen Tien Cuong$^2$ \quad Nguyen Thi Thu Trang$^{1^{*}}$\thanks{*Corresponding author.}
\end{tabular}
}
\address{
\begin{tabular}{c}
$^1$Hanoi University of Science and Technology \quad
$^2$VNPT AI, VNPT Group \\
$^3$National Economics University
\end{tabular}
}

\begin{document}
%
\maketitle
\begin{abstract}
Mispronunciation Detection and Diagnosis (MDD) is crucial for language learning and speech therapy. Unlike conventional methods that require scoring models or training phoneme-level models, we propose a novel training-free framework that leverages retrieval techniques with a pretrained Automatic Speech Recognition model. Our method avoids phoneme-specific modeling or additional task-specific training, while still achieving accurate detection and diagnosis of pronunciation errors. Experiments on the L2-ARCTIC dataset show that our method achieves a superior F1 score of 69.60\% while avoiding the complexity of model training.
\end{abstract}
\section{Introduction}
\label{sec:intro}

Mispronunciation Detection and Diagnosis is a fundamental task in Computer-Assisted Pronunciation Training (CAPT). Given a reference text and a learner’s spoken utterance, an MDD system must determine whether the pronunciation is correct, localize mispronounced units, and provide diagnostic feedback. The earliest systems were built on the Goodness-of-Pronunciation (GOP), which use acoustic models to compute phoneme scores and use a threshold to detect mispronunciation \cite{gop1}. While GOP offers a simple and efficient approach for scoring pronunciation accuracy, it lacks detailed diagnostic feedback for learners.

With the rise of end-to-end Automatic Speech Recognition (ASR), MDD research has shifted toward phoneme recognition models. The phoneme sequences recognized from the learner's audio by these models are aligned with the canonical phoneme sequence from the reference text to perform detection and diagnosis. One notable ASR-based model is CNN-RNN-CTC architecture \cite{cnn-rnn-ctc}. This model processes input audio through a deep learning framework trained with the Connectionist Temporal Classification (CTC) \cite{ctc} loss function, directly mapping raw acoustic features to pronounced phoneme sequences. While it achieves superior results over former methods, this approach does not use prior text information that learners were expected to read aloud.
 
To address this limitation, \cite{sed-mdd} proposed an end-to-end architecture for sentence-dependent MDD, combining acoustic features with a predefined canonical text sequence via cross-attention \cite{attn}. However, mismatches between the reference text and predicted phoneme sequence often cause inconsistencies. To overcome this issue, Fu et al. \cite{AFullText-Dependent} replaced the reference text with canonical phoneme sequences and applied data augmentation to improve input diversity and robustness.
 
Pretrained self-supervised speech models and ASR models have become widely used in MDD systems \cite{APL,related_peng21e_interspeech}. By leveraging large-scale unlabeled or transcribed speech corpora, these models can encode raw audio into robust phonetic embeddings through transfer learning, providing a rich representation of speech. Such embeddings significantly improve the ability of MDD systems. In parallel, specially designed linguistic encoders, such as text-aware and graph-based methods, have been explored to enhance the linguistic branch \cite{gcnmdd, text-aware}. These approaches capture structured dependencies among canonical phonemes, phonological rules, and articulatory attributes, enabling MDD systems to incorporate prior linguistic knowledge more effectively.

More recently, multitask approaches has gained considerable attention in MDD. Instead of relying solely on the phoneme recognition, \cite{multilingualmdd} proposed a multilingual MDD framework that leverages the speaker’s native language (L1) and target language (L2) information. This design explicitly models cross-lingual phonological disparities, thereby improving detection robustness. In addition, several studies have shown the benefits of joint training for MDD and Automatic Pronunciation Assessment, demonstrating that the two tasks are highly correlated and that their integration leads to more reliable error detection and diagnosis \cite{join_mdd_apa}. Other works have explored multi-view speech representation approaches, where multiple pretrained audio encoders are fused to strengthen phonetic representations, achieving improved generalization across diverse groups of L2 learners \cite{multiview}.

In this work, we find that modern ASR systems already encode sufficient linguistic information to support mispronunciation detection. Therefore, inspired by retrieval-based methods, we discover that MDD can be performed without phoneme-specific modeling or additional task-specific training, while still providing the necessary detail for accurate detection and diagnosis of pronunciation errors. Particularly, our contributions are summarized as follows:
\begin{enumerate}
\item We propose a training-free MDD framework that eliminates the need for phoneme-level modeling.
\item We introduce the first retrieval-based strategy for MDD, leveraging pretrained ASR models in a RAG-inspired manner.
\item We show through experiments on public L2 English datasets that our approach achieves competitive detection accuracy with minimal complexity.
\end{enumerate}


\section{Proposed Method}
\label{sec:proposal}

\begin{figure}[h!]
  \centering
    \includegraphics[width=1.07\linewidth]{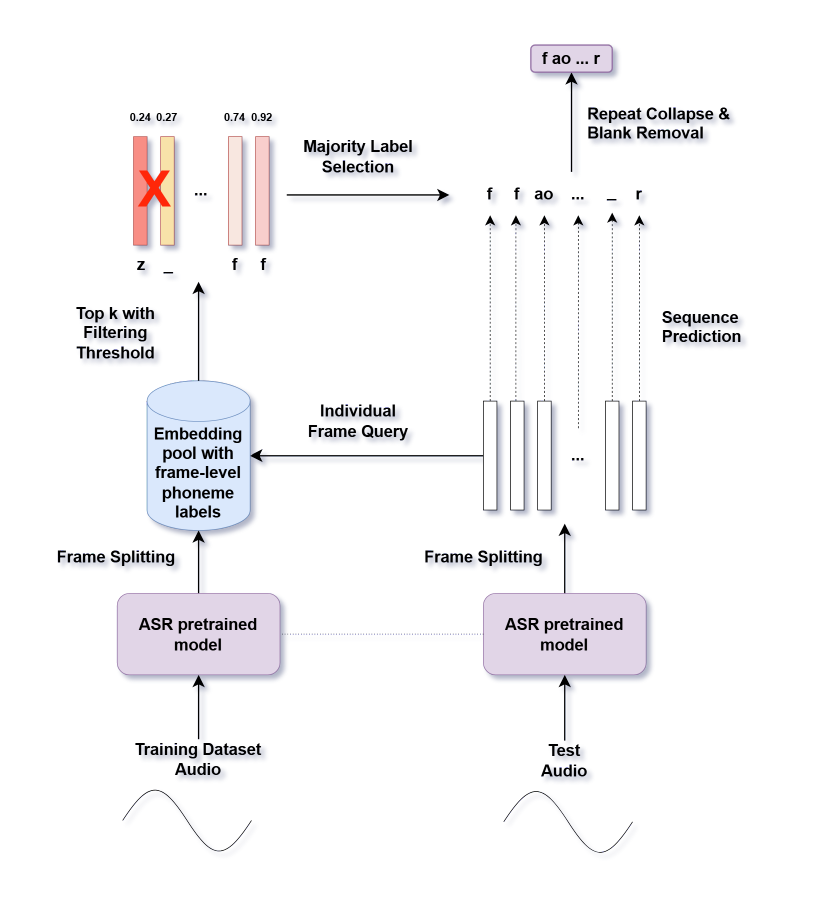}
    \caption{Illustration of our proposed PER-MDD method.}
    \label{fig:pipeline}
\end{figure}

Retrieval-based strategies have recently demonstrated strong effectiveness across a variety of domains, including enhancing large language models \cite{rag-llm} and speech processing \cite{asr-retrival,vc-retrival}
. In this paper, we propose a retrieval-based pipeline for MDD, which is named Phoneme Embedding Retrieval MDD (PER-MDD). The overall pipeline illustrates in figure \ref{fig:pipeline}.

\subsection{Phoneme embedding pool construction}

Let $\mathcal{D}_{\text{train}}$ be a labeled training dataset where each utterance has phoneme-level time alignments.
Each utterance is segmented into frames $\{x_t\}_{t=1}^T$, and a pretrained ASR model $f(\cdot)$ maps each frame to an embedding:
\begin{equation}
e_t = f(x_t), \quad e_t \in R^d.
\end{equation}

Since each frame corresponds to a specific time in the audio, and the dataset provides phoneme labels with start and end times, we can assign each frame a phoneme label $y_t \in \mathcal{V} \cup {\text{blank}}$, where $\mathcal{V}$ is the phoneme vocabulary.

The phoneme embedding pool is then defined as:
\begin{equation}
\mathcal{P} = {(e_t, y_t)}_{t=1}^N,
\end{equation}
with $N$ total frame-level pairs.

To construct the pool, we can select embeddings in different ways for each phoneme span (from start time $s$ to end time $e$ of embedding audio with phoneme labels):

\begin{itemize}
    \item \textbf{All-frame:} include every frame embedding $e_t$.
    \item \textbf{Middle-frame:} include only the embedding of the middle frame of the span, $e_{\lfloor (s+e)/2 \rfloor}$.
    \item \textbf{Mean-frame:} compute the average embedding over the span $\bar{e} = \frac{1}{e-s}\sum_{t=s}^e e_t$.
\end{itemize}

This pool $\mathcal{P}$ serves as the reference for retrieval-based phoneme prediction during inference.

\subsection{Inference procedure}

For a test utterance, we extract frame-level embeddings $q_t = f(x_t^{\text{test}})$.  
Each query $q_t$ is compared against the pool $\mathcal{P}$ using cosine similarity:
\begin{equation}
s(q_t, e) = \frac{q_t \cdot e}{\|q_t\|\|e\|}.
\end{equation}

\noindent\textbf{Candidate Retrieval.}
The top-$k$ nearest neighbors are selected:
\begin{equation}
\mathcal{N}_k(q_t) = \text{Top-}k\big(\{ (e,y) \in \mathcal{P} \mid s(q_t, e) \}\big).
\end{equation}
Then, a filtering threshold $\tau$ is applied:
\begin{equation}
\mathcal{N}^\ast(q_t) = \{ y \mid (e,y) \in \mathcal{N}_k(q_t), \; s(q_t,e) \geq \tau \}.
\end{equation}

\noindent\textbf{Label Assignment.}  
The predicted label at frame level is:
\begin{equation}
\hat{y}_t =
\begin{cases}
\text{blank}, & \text{if } \mathcal{N}^\ast(q_t) = \emptyset, \\
\mathrm{mode}(\mathcal{N}^\ast(q_t)), & \text{otherwise.}
\end{cases}
\end{equation}

\begin{table*}
\caption{Performance comparison between our proposed model and its baselines}
\scalebox{1.05}{
\begin{tabular}{|l|ccccccc|cc|}
\hline
\multirow{2}{*}{\textbf{Model}} & \multicolumn{7}{c|}{\textbf{MDD Metric}}                                                                                                                                                                                                                                                                       & \multicolumn{2}{c|}{\textbf{ASR metric}}                     \\ \cline{2-10} 
                                & \multicolumn{1}{c|}{\textbf{FRR$\downarrow$}} & \multicolumn{1}{c|}{\textbf{FAR$\downarrow$}} & \multicolumn{1}{c|}{\textbf{DER$\downarrow$}} & \multicolumn{1}{c|}{\textbf{PRE$\uparrow$}} & \multicolumn{1}{c|}{\textbf{REC$\uparrow$}} & \multicolumn{1}{c|}{\textbf{F1$\uparrow$}} & \textbf{DA$\uparrow$} & \multicolumn{1}{c|}{\textbf{PER$\downarrow$}} & \textbf{COR$\uparrow$} \\ \hline
PHN-M2 \cite{slat2025}                         & \multicolumn{1}{c|}{6.33}                     & \multicolumn{1}{c|}{45.37}                    & \multicolumn{1}{c|}{\textbf{25.12}}           & \multicolumn{1}{c|}{64.51}                  & \multicolumn{1}{c|}{54.63}                  & \multicolumn{1}{c|}{59.16}                 & 86.88                 & \multicolumn{1}{c|}{17.12}                    & -            \\ \hline
L1-MultiMDD \cite{multilingualmdd}                    & \multicolumn{1}{c|}{4.60}                     & \multicolumn{1}{c|}{-}                        & \multicolumn{1}{c|}{-}                        & \multicolumn{1}{c|}{-}                      & \multicolumn{1}{c|}{-}                      & \multicolumn{1}{c|}{57.40}                 & -                     & \multicolumn{1}{c|}{\textbf{12.55}}           & -            \\ \hline
w2v2-XLSR \cite{related_peng21e_interspeech}                     & \multicolumn{1}{c|}{5.70}                     & \multicolumn{1}{c|}{41.80}                    & \multicolumn{1}{c|}{29.28}                    & \multicolumn{1}{c|}{62.86}                  & \multicolumn{1}{c|}{58.20}                  & \multicolumn{1}{c|}{60.44}                 & -                     & \multicolumn{1}{c|}{16.20}                    & -            \\ \hline
Joint-Align \cite{joint-align}                     & \multicolumn{1}{c|}{-}                        & \multicolumn{1}{c|}{-}                        & \multicolumn{1}{c|}{-}                        & \multicolumn{1}{c|}{\textbf{77.12}}         & \multicolumn{1}{c|}{53.31}                  & \multicolumn{1}{c|}{63.04}                 & -                     & \multicolumn{1}{c|}{-}                        & -            \\ \hline
MDDGCN \cite{gcnmdd}                         & \multicolumn{1}{c|}{9.18}                     & \multicolumn{1}{c|}{38.03}                    & \multicolumn{1}{c|}{25.24}                    & \multicolumn{1}{c|}{51.90}                  & \multicolumn{1}{c|}{61.97}                  & \multicolumn{1}{c|}{56.49}                 & -                     & \multicolumn{1}{c|}{-}                        & -            \\ \hline
MVmulti−MTseq \cite{multiview}                   & \multicolumn{1}{c|}{-}                        & \multicolumn{1}{c|}{-}                        & \multicolumn{1}{c|}{-}                        & \multicolumn{1}{c|}{61.43}                  & \multicolumn{1}{c|}{59.23}                  & \multicolumn{1}{c|}{60.31}                 & -                     & \multicolumn{1}{c|}{14.13}                    & -            \\ \hline
PER-MDD (Ours)                  & \multicolumn{1}{c|}{\textbf{4.43}}            & \multicolumn{1}{c|}{\textbf{32.44}}           & \multicolumn{1}{c|}{37.77}                    & \multicolumn{1}{c|}{71.78}                  & \multicolumn{1}{c|}{\textbf{67.56}}         & \multicolumn{1}{c|}{\textbf{69.60}}        & \textbf{91.57}        & \multicolumn{1}{c|}{104.08}                   & 90.42        \\ \hline
\end{tabular}
}
\label{main-result}
\\
\centering
{\small $\downarrow$ lower is better}
{\small $\uparrow$ higher is better}
\end{table*}

\noindent\textbf{Post-processing.}  
The sequence $\{\hat{y}_t\}$ is refined by:  
(i) collapsing consecutive duplicates, and  
(ii) removing blanks.  
This produces a final phoneme sequence $\hat{Y}$ that represents the predicted pronunciation, which is aligned with the canonical phoneme sequence to detect and diagnose mispronunciations.

\section{Experiments}

\subsection{Datasets}

We evaluate our model's performance using the publicly available L2-ARCTIC dataset \cite{L2arctic}, which contains non-native English speech from speakers of various native languages, including Hindi, Korean, Mandarin, Spanish, Arabic, and Vietnamese (24 speakers in total). L2-ARCTIC is specifically designed for CAPT tasks and provides canonical phoneme sequences from the reference text, corresponding audio recordings, and the actual phonemes produced by each speaker. Following prior work \cite{AFullText-Dependent, sed-mdd, joint-align}, we use six speakers (“NJS”, “TLV”, “TNI”, “TXHC”, “YKWK”, “ZHAA”) to construct the test set, and 12 speakers to build the training set.

\subsection{Experimental setup}\label{sub:exp-setup}
We use the publicly available HuBERT model \cite{hubert}, fine-tuned on a large-scale ASR dataset\footnote{\url{https://huggingface.co/facebook/hubert-large-ls960-ft}}
. We sample 500 training audio files to build the phoneme embedding pool, representing each span with its middle frame (known as mid-frame pooling). The similarity threshold is 0.7, and the retrieval top-$k$ is 10.

\begin{table}[h!]
\centering
\caption{MDD evaluation example}
\scalebox{0.95}{
\begin{tabular}{|l|c|c|c|c|c|}
\hline
Canonical phoneme & ae & d & v & ay & s\\ \hline
Human-annotated phoneme & {\color[HTML]{32CB00} ae} & {\color[HTML]{FE0000} t} & {\color[HTML]{32CB00} v} & {\color[HTML]{FE0000} ey} & {\color[HTML]{FE0000} sh} \\ \hline
Predicted phoneme & {\color[HTML]{32CB00} ae} & {\color[HTML]{32CB00} d} & {\color[HTML]{FE0000} f} & {\color[HTML]{FE0000} ey} & {\color[HTML]{FE0000} z} \\ \hline
Evaluation result & TA & FA & FR & \begin{tabular}[c]{@{}c@{}}TR\\ CD\end{tabular} & \begin{tabular}[c]{@{}c@{}}TR\\ DE\end{tabular} \\ \hline
\end{tabular}
}
\label{table:eval}
\end{table}
\subsection{Evaluation metrics}
Consistent with previous studies \cite{sed-mdd,cnn-rnn-ctc,AFullText-Dependent,APL}, we evaluate our models using both ASR and MDD performance metrics. For ASR, we use the phone error rate (PER) and Correctness (COR) to assess performance. The MDD evaluation process involves categorizing the model's predictions into distinct groups: true acceptance (TA), true rejection (TR), false acceptance (FA), and false rejection (FR). Within TR, we further divide results into correct diagnosis (CD) and error diagnosis (DE). Metrics such as detection accuracy (DA), diagnosis error rate (DER), recall (REC), precision (PRE), and F1 score are computed to assess the model's performance. An illustrative example of the MDD evaluation is presented in Table \ref{table:eval}. A green phoneme is one that is the same as the canonical phoneme, while a red phoneme is one that is different. For more details on how all the metrics are computed, refer to \cite{APL}.

\subsection{Performance analysis}
\subsubsection{Performance comparison with baslines}
In MDD systems, misjudging a large number of correctly pronounced phones as mispronunciations can frustrate learners and negatively impact their learning experience. Therefore, FRR is regarded as the most critical metric in MDD. As shown in Table \ref{main-result}, our proposed model achieves superior performance with an FRR of only 4.43\%. Moreover, our model also delivers significant improvements in F1, FAR, and recall, providing significant performance gains of around 6\%, compared to state-of-the-art (SOTA) baselines such as MDDGCN \cite{gcnmdd} and Joint-Align \cite{joint-align}. Note from this table that since some baseline models are not open-source, we report the official results released by the authors. The parameters of our scheme shown in Table \ref{main-result} are provided in Subsection~\ref{sub:exp-setup}.

\begin{table*}
\caption{Ablation studies on our PER-MDD}
\begin{tabular}{|l|c|l|c|l|c|c|c|c|c|c|}
\hline
\rowcolor[HTML]{CBCEFB} 
{\color[HTML]{000000} \textbf{ASR model}}                               & {\color[HTML]{000000} \textbf{Top-k}}                                & \multicolumn{1}{c|}{\cellcolor[HTML]{CBCEFB}{\color[HTML]{000000} \textbf{Pool size}}}    & {\color[HTML]{000000} \textbf{Threshold}}                            & \multicolumn{1}{c|}{\cellcolor[HTML]{CBCEFB}{\color[HTML]{000000} \textbf{Strategy}}}     & {\color[HTML]{000000} \textbf{PER$\downarrow$}} & {\color[HTML]{000000} \textbf{REC$\uparrow$}} & {\color[HTML]{000000} \textbf{PRE$\uparrow$}} & {\color[HTML]{000000} \textbf{F1$\uparrow$}} & {\color[HTML]{000000} \textbf{FRR$\downarrow$}} & {\color[HTML]{000000} \textbf{DER$\downarrow$}} \\ \hline
\rowcolor[HTML]{FFFFFF} 
{\color[HTML]{000000} Data2vec}                                         & \multicolumn{1}{l|}{\cellcolor[HTML]{FFFFFF}{\color[HTML]{000000} }} & {\color[HTML]{000000} }                                                                   & \multicolumn{1}{l|}{\cellcolor[HTML]{FFFFFF}{\color[HTML]{000000} }} & {\color[HTML]{000000} }                                                                   & {\color[HTML]{000000} 188.20}                   & {\color[HTML]{000000} \textbf{75.25}}         & {\color[HTML]{000000} 55.57}                  & {\color[HTML]{000000} 63.93}                 & {\color[HTML]{000000} 10.04}                    & {\color[HTML]{000000} 47.85}                    \\
\rowcolor[HTML]{FFFFFF} 
{\color[HTML]{000000} Wav2vec2}                                         & {\color[HTML]{000000} 10}                                            & \multicolumn{1}{c|}{\cellcolor[HTML]{FFFFFF}{\color[HTML]{000000} 500}}                   & {\color[HTML]{000000} 0.7}                                           & \multicolumn{1}{c|}{\cellcolor[HTML]{FFFFFF}{\color[HTML]{000000} Mid}}                   & {\color[HTML]{000000} 173.28}                   & {\color[HTML]{000000} 74.46}                  & {\color[HTML]{000000} 64.58}                  & {\color[HTML]{000000} 69.17}                 & {\color[HTML]{000000} 6.81}                     & {\color[HTML]{000000} 43.54}                    \\
\rowcolor[HTML]{FFFFFF} 
{\color[HTML]{000000} Hubert}                                           & \multicolumn{1}{l|}{\cellcolor[HTML]{FFFFFF}{\color[HTML]{000000} }} & {\color[HTML]{000000} }                                                                   & \multicolumn{1}{l|}{\cellcolor[HTML]{FFFFFF}{\color[HTML]{000000} }} & {\color[HTML]{000000} }                                                                   & {\color[HTML]{000000} \textbf{104.08}}          & {\color[HTML]{000000} 67.56}                  & {\color[HTML]{000000} \textbf{71.78}}         & {\color[HTML]{000000} \textbf{69.60}}        & {\color[HTML]{000000} \textbf{4.43}}            & {\color[HTML]{000000} \textbf{37.77}}           \\ \hline
\rowcolor[HTML]{CBCEFB} 
{\color[HTML]{000000} }                                                 & {\color[HTML]{000000} 5}                                             & {\color[HTML]{000000} }                                                                   & \multicolumn{1}{l|}{\cellcolor[HTML]{CBCEFB}{\color[HTML]{000000} }} & {\color[HTML]{000000} }                                                                   & {\color[HTML]{000000} 135.27}                   & {\color[HTML]{000000} \textbf{71.01}}         & {\color[HTML]{000000} 72.39}                  & {\color[HTML]{000000} \textbf{71.69}}        & {\color[HTML]{000000} 4.52}                     & {\color[HTML]{000000} 39.19}                    \\
\rowcolor[HTML]{CBCEFB} 
{\color[HTML]{000000} }                                                 & {\color[HTML]{000000} 6}                                             & {\color[HTML]{000000} }                                                                   & \multicolumn{1}{l|}{\cellcolor[HTML]{CBCEFB}{\color[HTML]{000000} }} & {\color[HTML]{000000} }                                                                   & {\color[HTML]{000000} 123.36}                   & {\color[HTML]{000000} 69.59}                  & {\color[HTML]{000000} \textbf{72.71}}         & {\color[HTML]{000000} 71.11}                 & {\color[HTML]{000000} \textbf{4.36}}            & {\color[HTML]{000000} 38.31}                    \\
\rowcolor[HTML]{CBCEFB} 
\cellcolor[HTML]{CBCEFB}{\color[HTML]{000000} }                         & {\color[HTML]{000000} 7}                                             & \multicolumn{1}{c|}{\cellcolor[HTML]{CBCEFB}{\color[HTML]{000000} }}                      & \cellcolor[HTML]{CBCEFB}{\color[HTML]{000000} }                      & \multicolumn{1}{c|}{\cellcolor[HTML]{CBCEFB}{\color[HTML]{000000} }}                      & {\color[HTML]{000000} 116.74}                   & {\color[HTML]{000000} 69.10}                  & {\color[HTML]{000000} 72.56}                  & {\color[HTML]{000000} 70.79}                 & {\color[HTML]{000000} \textbf{4.36}}            & {\color[HTML]{000000} 38.28}                    \\
\rowcolor[HTML]{CBCEFB} 
\multirow{-2}{*}{\cellcolor[HTML]{CBCEFB}{\color[HTML]{000000} Hubert}} & {\color[HTML]{000000} 8}                                             & \multicolumn{1}{c|}{\multirow{-2}{*}{\cellcolor[HTML]{CBCEFB}{\color[HTML]{000000} 500}}} & \multirow{-2}{*}{\cellcolor[HTML]{CBCEFB}{\color[HTML]{000000} 0.7}} & \multicolumn{1}{c|}{\multirow{-2}{*}{\cellcolor[HTML]{CBCEFB}{\color[HTML]{000000} Mid}}} & {\color[HTML]{000000} 112.36}                   & {\color[HTML]{000000} 69.03}                  & {\color[HTML]{000000} 72.53}                  & {\color[HTML]{000000} 70.73}                 & {\color[HTML]{000000} \textbf{4.36}}            & {\color[HTML]{000000} 37.81}                    \\
\rowcolor[HTML]{CBCEFB} 
{\color[HTML]{000000} }                                                 & {\color[HTML]{000000} 9}                                             & {\color[HTML]{000000} }                                                                   & \multicolumn{1}{l|}{\cellcolor[HTML]{CBCEFB}{\color[HTML]{000000} }} & {\color[HTML]{000000} }                                                                   & {\color[HTML]{000000} 107.36}                   & {\color[HTML]{000000} 68.21}                  & {\color[HTML]{000000} 71.67}                  & {\color[HTML]{000000} 69.90}                 & {\color[HTML]{000000} 4.50}                     & {\color[HTML]{000000} \textbf{37.38}}           \\
\rowcolor[HTML]{CBCEFB} 
{\color[HTML]{000000} }                                                 & {\color[HTML]{000000} 10}                                            & {\color[HTML]{000000} }                                                                   & \multicolumn{1}{l|}{\cellcolor[HTML]{CBCEFB}{\color[HTML]{000000} }} & {\color[HTML]{000000} }                                                                   & {\color[HTML]{000000} \textbf{104.08}}          & {\color[HTML]{000000} 67.56}                  & {\color[HTML]{000000} 71.78}                  & {\color[HTML]{000000} 69.60}                 & {\color[HTML]{000000} 4.43}                     & {\color[HTML]{000000} 37.77}                    \\ \hline
\rowcolor[HTML]{FFFFFF} 
{\color[HTML]{000000} }                                                 & \multicolumn{1}{l|}{\cellcolor[HTML]{FFFFFF}{\color[HTML]{000000} }} & \multicolumn{1}{c|}{\cellcolor[HTML]{FFFFFF}{\color[HTML]{000000} 100}}                   & \multicolumn{1}{l|}{\cellcolor[HTML]{FFFFFF}{\color[HTML]{000000} }} & {\color[HTML]{000000} }                                                                   & {\color[HTML]{000000} 140.12}                   & {\color[HTML]{000000} \textbf{73.55}}         & {\color[HTML]{000000} 57.38}                  & {\color[HTML]{000000} 64.47}                 & {\color[HTML]{000000} 9.11}                     & {\color[HTML]{000000} 40.94}                    \\
\rowcolor[HTML]{FFFFFF} 
\cellcolor[HTML]{FFFFFF}{\color[HTML]{000000} }                         & \cellcolor[HTML]{FFFFFF}{\color[HTML]{000000} }                      & \multicolumn{1}{c|}{\cellcolor[HTML]{FFFFFF}{\color[HTML]{000000} 200}}                   & \cellcolor[HTML]{FFFFFF}{\color[HTML]{000000} }                      & \multicolumn{1}{c|}{\cellcolor[HTML]{FFFFFF}{\color[HTML]{000000} }}                      & {\color[HTML]{000000} 123.49}                   & {\color[HTML]{000000} 71.20}                  & {\color[HTML]{000000} 65.28}                  & {\color[HTML]{000000} 68.11}                 & {\color[HTML]{000000} 6.32}                     & {\color[HTML]{000000} 40.39}                    \\
\rowcolor[HTML]{FFFFFF} 
\multirow{-2}{*}{\cellcolor[HTML]{FFFFFF}{\color[HTML]{000000} Hubert}} & \multirow{-2}{*}{\cellcolor[HTML]{FFFFFF}{\color[HTML]{000000} 10}}  & \multicolumn{1}{c|}{\cellcolor[HTML]{FFFFFF}{\color[HTML]{000000} 500}}                   & \multirow{-2}{*}{\cellcolor[HTML]{FFFFFF}{\color[HTML]{000000} 0.7}} & \multicolumn{1}{c|}{\multirow{-2}{*}{\cellcolor[HTML]{FFFFFF}{\color[HTML]{000000} Mid}}} & {\color[HTML]{000000} 104.08}                   & {\color[HTML]{000000} 67.56}                  & {\color[HTML]{000000} 71.78}                  & {\color[HTML]{000000} \textbf{69.60}}        & {\color[HTML]{000000} 4.43}                     & {\color[HTML]{000000} 37.77}                    \\
\rowcolor[HTML]{FFFFFF} 
{\color[HTML]{000000} }                                                 & \multicolumn{1}{l|}{\cellcolor[HTML]{FFFFFF}{\color[HTML]{000000} }} & \multicolumn{1}{c|}{\cellcolor[HTML]{FFFFFF}{\color[HTML]{000000} 1800}}                  & \multicolumn{1}{l|}{\cellcolor[HTML]{FFFFFF}{\color[HTML]{000000} }} & {\color[HTML]{000000} }                                                                   & {\color[HTML]{000000} \textbf{84.63}}           & {\color[HTML]{000000} 62.83}                  & {\color[HTML]{000000} \textbf{77.49}}         & {\color[HTML]{000000} 69.40}                 & {\color[HTML]{000000} \textbf{3.04}}            & {\color[HTML]{000000} \textbf{36.91}}           \\ \hline
\rowcolor[HTML]{CBCEFB} 
{\color[HTML]{000000} }                                                 & \multicolumn{1}{l|}{\cellcolor[HTML]{CBCEFB}{\color[HTML]{000000} }} & {\color[HTML]{000000} }                                                                   & {\color[HTML]{000000} No}                                            & {\color[HTML]{000000} }                                                                   & {\color[HTML]{000000} 102.10}                   & {\color[HTML]{000000} 67.65}                  & {\color[HTML]{000000} 71.64}                  & {\color[HTML]{000000} 69.59}                 & {\color[HTML]{000000} 4.47}                     & {\color[HTML]{000000} \textbf{37.75}}           \\
\rowcolor[HTML]{CBCEFB} 
{\color[HTML]{000000} }                                                 & \multicolumn{1}{l|}{\cellcolor[HTML]{CBCEFB}{\color[HTML]{000000} }} & {\color[HTML]{000000} }                                                                   & {\color[HTML]{000000} 0.6}                                           & {\color[HTML]{000000} }                                                                   & {\color[HTML]{000000} 102.32}                   & {\color[HTML]{000000} 67.63}                  & {\color[HTML]{000000} 71.74}                  & {\color[HTML]{000000} \textbf{69.63}}        & {\color[HTML]{000000} 4.44}                     & {\color[HTML]{000000} 37.77}                    \\
\rowcolor[HTML]{CBCEFB} 
{\color[HTML]{000000} Hubert}                                           & {\color[HTML]{000000} 10}                                            & \multicolumn{1}{c|}{\cellcolor[HTML]{CBCEFB}{\color[HTML]{000000} 500}}                   & {\color[HTML]{000000} 0.7}                                           & \multicolumn{1}{c|}{\cellcolor[HTML]{CBCEFB}{\color[HTML]{000000} Mid}}                   & {\color[HTML]{000000} 104.08}                   & {\color[HTML]{000000} 67.56}                  & {\color[HTML]{000000} \textbf{71.78}}         & {\color[HTML]{000000} 69.60}                 & {\color[HTML]{000000} \textbf{4.43}}            & {\color[HTML]{000000} 37.77}                    \\
\rowcolor[HTML]{CBCEFB} 
{\color[HTML]{000000} }                                                 & \multicolumn{1}{l|}{\cellcolor[HTML]{CBCEFB}{\color[HTML]{000000} }} & {\color[HTML]{000000} }                                                                   & {\color[HTML]{000000} 0.8}                                           & {\color[HTML]{000000} }                                                                   & {\color[HTML]{000000} 101.90}                   & {\color[HTML]{000000} 69.28}                  & {\color[HTML]{000000} 67.23}                  & {\color[HTML]{000000} 68.24}                 & {\color[HTML]{000000} 5.63}                     & {\color[HTML]{000000} 41.47}                    \\
\rowcolor[HTML]{CBCEFB} 
{\color[HTML]{000000} }                                                 & \multicolumn{1}{l|}{\cellcolor[HTML]{CBCEFB}{\color[HTML]{000000} }} & {\color[HTML]{000000} }                                                                   & {\color[HTML]{000000} 0.9}                                           & {\color[HTML]{000000} }                                                                   & {\color[HTML]{000000} \textbf{76.60}}           & {\color[HTML]{000000} \textbf{73.18}}         & {\color[HTML]{000000} 37.14}                  & {\color[HTML]{000000} 49.27}                 & {\color[HTML]{000000} 20.66}                    & {\color[HTML]{000000} 55.16}                    \\ \hline
\rowcolor[HTML]{FFFFFF} 
{\color[HTML]{000000} }                                                 & \multicolumn{1}{l|}{\cellcolor[HTML]{FFFFFF}{\color[HTML]{000000} }} & {\color[HTML]{000000} }                                                                   & \multicolumn{1}{l|}{\cellcolor[HTML]{FFFFFF}{\color[HTML]{000000} }} & \multicolumn{1}{c|}{\cellcolor[HTML]{FFFFFF}{\color[HTML]{000000} All}}                   & {\color[HTML]{000000} \textbf{76.17}}           & {\color[HTML]{000000} 61.90}                  & {\color[HTML]{000000} 63.94}                  & {\color[HTML]{000000} 62.90}                 & {\color[HTML]{000000} 5.82}                     & {\color[HTML]{000000} 41.98}                    \\
\rowcolor[HTML]{FFFFFF} 
{\color[HTML]{000000} Hubert}                                           & {\color[HTML]{000000} 10}                                            & \multicolumn{1}{c|}{\cellcolor[HTML]{FFFFFF}{\color[HTML]{000000} 500}}                   & {\color[HTML]{000000} 0.7}                                           & \multicolumn{1}{c|}{\cellcolor[HTML]{FFFFFF}{\color[HTML]{000000} Mean}}                  & {\color[HTML]{000000} 101.02}                   & {\color[HTML]{000000} 65.81}                  & {\color[HTML]{000000} 67.43}                  & {\color[HTML]{000000} 66.61}                 & {\color[HTML]{000000} 5.30}                     & {\color[HTML]{000000} 42.81}                    \\
\rowcolor[HTML]{FFFFFF} 
{\color[HTML]{000000} }                                                 & \multicolumn{1}{l|}{\cellcolor[HTML]{FFFFFF}{\color[HTML]{000000} }} & {\color[HTML]{000000} }                                                                   & \multicolumn{1}{l|}{\cellcolor[HTML]{FFFFFF}{\color[HTML]{000000} }} & \multicolumn{1}{c|}{\cellcolor[HTML]{FFFFFF}{\color[HTML]{000000} Mid}}                   & {\color[HTML]{000000} 104.08}                   & {\color[HTML]{000000} \textbf{67.56}}         & {\color[HTML]{000000} \textbf{71.78}}         & {\color[HTML]{000000} \textbf{69.60}}        & {\color[HTML]{000000} \textbf{4.43}}            & {\color[HTML]{000000} \textbf{37.77}}           \\ \hline
\end{tabular}
\label{ablation:result}
\\
\centering
{\small $\downarrow$ lower is better}
{\small $\uparrow$ higher is better}
\end{table*}

It is worth noting from Table \ref{main-result}  that our retrieval-based MDD method tends to produce a relatively large number of insertion errors compared to human-annotated transcripts, leading to considerably higher PER than baselines. However, these insertion errors do not significantly affect MDD performance. As described in the metrics of \cite{cnn-rnn-ctc}, MDD systems primarily evaluate whether each phoneme in the canonical phoneme sequence is pronounced correctly, focusing on diagnosing errors such as whether a phoneme is pronounced as another, rather than detecting insertions.

To further illustrate this, the Correctness, computed similarly to ($1-\text{PER}$) but considering only substitution and deletion errors, reaches 90.42\%, indicating that our model predicts phonemes largely in agreement with human annotations.

\subsubsection{Ablation study}
We conduct an ablation study by modifying several key components to evaluate their individual contributions to the performance of our PER-MDD, as summarized in Table \ref{ablation:result}. 

First, we examine the effect of different ASR models on performance by comparing the default HuBERT configuration with two additional models: Data2vec \cite{data2vec}\footnote{\url{https://huggingface.co/facebook/data2vec-audio-large-960h}}
 and Wav2vec2 \cite{wav2vec2}\footnote{\url{https://huggingface.co/facebook/wav2vec2-large-960h-lv60}}. Among them, HuBERT achieves the best performance on this task. 

Next, we vary the threshold to quantify its influence on performance. The results show only minor differences between no threshold and a threshold of 0.7, as most top-$k$ candidates already have cosine similarity scores above this level. When the threshold is set higher, at 0.9, PER decreases but F1 is considerably reduced, indicating a necessary trade-off.

We then explore the impact of the top-$k$ value. Decresing $k$ generally worsens PER, but MDD performance shows slight improvements, suggesting a balance between retrieval precision and coverage. 

We also evaluate different pooling strategies for constructing the vector database, including all-frame pooling, mean-frame pooling, and mid-frame pooling. Among them, the mid-frame strategy achieves the best performance. In addition to better results, it also reduces the pool size, leading to faster query times.

Finally, we vary the pool size, using 100 random utterances, 200 utterances, 500 utterances, and the full set of 1,800 utterances from the training dataset. The results clearly show that larger pool sizes reduce insertion errors and improve MDD metrics overall.

\section{Conclusions}
This paper explores the use of a retrieval-based approach to enhance MDD systems. We proposed a novel framework that integrates an ASR model with a retrieval-based method, thereby eliminating the need to train an additional phoneme recognition model. Our approach achieves significant improvements in both FRR and F1 score, reaching an FRR of only 4.43\%, the best among compared methods, and delivering a 6.56\% F1 gain over SOTA models. These findings underscore the effectiveness of retrieval-based strategies for advancing MDD. However, our model still struggles with insertion errors, which lead to a relatively high PER. In future work, we aim to develop more robust mechanisms for handling insertions and to further optimize retrieval efficiency.

\vfill\pagebreak

\bibliographystyle{IEEEbib}
\bibliography{strings}

\end{document}